\documentclass[letterpaper, 10 pt, conference]{ieeeconf}  % Comment this line out if you need a4paper

\IEEEoverridecommandlockouts                              % This command is only needed if 
\usepackage{graphics} % for pdf, bitmapped graphics files
\usepackage{epsfig} % for postscript graphics files
\usepackage{mathptmx} % assumes new font selection scheme installed
\usepackage{times} % assumes new font selection scheme installed
\usepackage{amsmath} % assumes amsmath package installed
\usepackage{amssymb}  % assumes amsmath package installed

\title{\LARGE \bf
A novel pLSA based Traffic Signs Classification System
}

\author{Mrinal Haloi$^{1}$% <-this % stops a space
\thanks{*This work was Accepted in APMediaCast-2015, Bali, Indonesia}% <-this % stops a space
\thanks{$^{1}$Mrinal Haloi with Indian Institute of Technology, Guwahati. 
        {\tt\small h.mrinal@iitg.ernet.in or mrinal.haloi11@gmail.com }; Copyright 2015 by the author}%
}

\begin{document}

\maketitle
\thispagestyle{empty}
\pagestyle{empty}

\section*{\centering Abstract}
\textbf{
In this work we developed a novel and fast traffic sign recognition system, a very important part for advanced driver assistance system and for autonomous driving. Traffic signs play a very vital role in safe driving and avoiding accident. We have used image processing and topic discovery model pLSA to tackle this challenging multiclass classification problem. Our algorithm is consist of two parts, shape classification and sign classification for improved accuracy. For processing and representation of image we have used bag of features model with SIFT local descriptor. Where a visual vocabulary of size 300 words are formed using k-means codebook formation algorithm. We exploited the concept that every image is a collection of visual topics and images having same topics will belong to same category. Our algorithm is tested on German traffic sign recognition benchmark (GTSRB) and gives very promising result near to existing state of the art techniques.
}\\
\textbf{KeyWords}: ADAS, Traffic Signs Classification, pLSA.

\section{Introduction}
Traffic sign recognition system play a important role in autonomous driving environment and for advanced driver assistance systems. For driving safely and avoiding accident in daily life’s traffic signs are very important. A smart system that can help human driver by automatic recognition and classification and giving warning will ease driver task.
With the advancement of modern vehicles design and safety this have got a considerable attention. For a human driver because of fatigue, divergence of attention, occlusion of sign due to road obstruction and natural scenes, related to different problems may lead to miss some important traffic sign, which may result in severe accident. Also automatic recognition system will help in proper navigation and to follow traffic rules.
 There are several challenges involved in developing a complete traffic sign recognition system, because of occlusion of signs due to different background, specifically trees, building etc. old damaged signs, weather condition, viewpoint variation, also illuminant changes, in day and night etc. A complete traffic sign system consist of detection and classification a wide variety of signs. Classification of different class signs having same shape is always a difficult part, since there are very small difference in between lots of traffic signs.

\begin{figure}
  \centering
      \includegraphics[width=3.2in,height=1.7in]{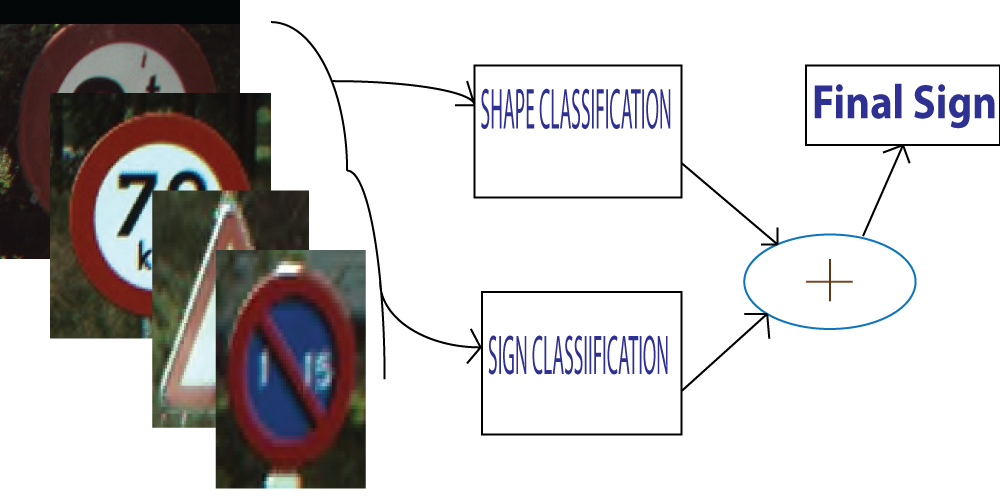}
\caption{Main Concept}
\end{figure}

In this work we will present a novel classification techniques based on probabilistic latent semantic analysis. Also we have built a shape classification system based of pyramid of HOG features and template matching. For feature representation of final pLSA system well known SIFT is used.

\begin{figure*}

 \center

  \includegraphics[width=5.4in, height = 2.8in]{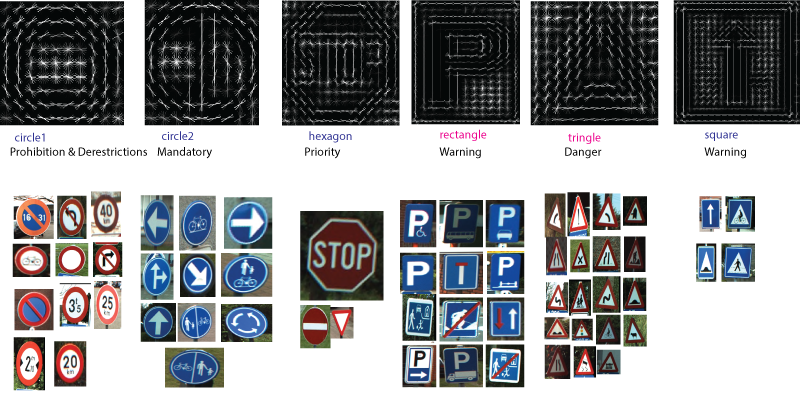}

  \caption{Sign category and its shapes}

  \label{AAA}

\end{figure*}

Previously different method which have got state of art result on GTSRB and GTSDB are very complex to train and need supervision. Method related to convolutional neural network, neural network, support vector machine are computationally complex and need effort to implement in system like FPGA device or other low computationally powerful devices. In our method related to pLSA is computationally flexible than previous method, and it is an unsupervised method and use KNN classifier at final step. 
 We will show our result on publicly available German traffic sign (GTSRB) database. We have an accuracy of around 96.86-100\% over sub category.
Rest of the paper is organised as follows, in section 2 we have reviewed existence literature in this area, section 3 describes our main algorithm and section 4 gives detail about dataset and experimental result along with comparison with other existing method.

\section{Related Work}
Considerable research work exist on detection and classification traffic signs for addressing the challenged involved in real life problems. Even though it is not possible go through all this research works in this paper we will give brief overview of some relevant works. Most of the work used computer vision and machine learning based techniques by using data from several camera sensors mounted on car roof at different angles. In some of the work researchers explores detection based on colour features, such as converting the colour space from RGB to HSV etc. and then using colour thresholding method for detection and classification by well-known support vector machines. In colour thresholding approach morphological operation like connected component analysis was done for accurate location. 
Colour, shape, motion information and haar wavelet based features was used in this work [12]. By using SVM based colour classification on a block of pixels Le et all [13] addressed the problems of weather variation. Features like, sift, hog and haar wavelet etc. was used by some of this work.
In German Traffic Sign Recognition Benchmark (GTSRB) competition, top performing algorithm exceeds best human classification accuracy. By using committee of neural networks [3] achieved highest ever performance of 99.46\%, where best human performance was 98.84\%. Multiscale convolutional network [15] has achieved 98.31\% accuracy in this dataset. Also other algorithm based on k-d trees and random forest [16] and LDA on HOG1 [2] have got very high accuracy.
For recognition purpose group sparse coding was used by [14] for rich feature learning of traffic signs recognition.

\section{Model for Classification}
The aim of the works is to develop a topic based classification frameworks for fast and reliable traffic sign categorization. Discovering hidden topics in images and correlating them with similar topics images form a successful classification algorithm. Since each of traffic sign category can be assumed very well as combination of one or more topics, so we have choose this method over other learning based method. 
For classification of the images we have used two step method, in first processing we will classify the shape of the traffic signs and after that we will classify its actual class. Image may have undergone through different type of rotational effect due to viewpoint variation, alignment. In Fig. [1] main idea concept of our method is depicted. As a prepossessing task we will use affine invariant transform on the images for getting rid of rotational effect.

\subsection{Shape Classification}
In this step traffic sign we will be divided into six class, specifically tringle, square, circle, single-circle, rectangle and hexagon as shown in Fig. [2]. Dividing the images in terms of shape help in classification since traffic sign topic most of the time depends on their shape. Different shape class images associated with different levels of dangers in road. Getting correct shape of the images is a boost for further classification. For shape representation HOG [7] is very successful. HOG accurately capture structures of an image, in low resolution of an image it capture high level features leaving fine grained pixels. This will help us representing the images as different skull shape and getting rid of its inside content. We have formed first five hog template for reference of five shapes. For classification a pyramid of hog feature at different resolution of the image is developed. Then a template matching based method is used for classification of its shape to a specific shape from the templates. This is based on normalization correlation value.

\begin{equation}
dist(T,Im) = \Sigma_{[i,j]\epsilon R}(T(i,j)-Im(i,j))^{2} 
\end{equation}

\subsection{Sign Classification}
In this step we have image of known shape, for classifying their actual sign on the basis of hidden topic, a probabilistic Latent Semantic Analysis based classification techniques will be used. Since image is a very high dimensional data, we will prepossess it to reduce its dimensionality by using visual codebook formation method. Each image will be represented as a bag of visual words. Each visual words will be relevant information stored in those images.

\subsubsection{Feature Extraction}
For extracting meaningful information from images we will use SIFT descriptor.
For object recognition SIFT is a very popular local descriptor having dimension of 128. This descriptor is scale invariant and suitable for object recognition and key-point matching purpose. SIFT computation based on construction of scale space. It uses difference of Gaussian filter for locating potential corner points. Actual implementation used 16x16 image patch, and further divided into 4x4 descriptors, resulting a 4x4x16 = 128 dimensional vector.

\begin{figure*}
 \center

  \includegraphics[width=5.4in, height = 2.3in]{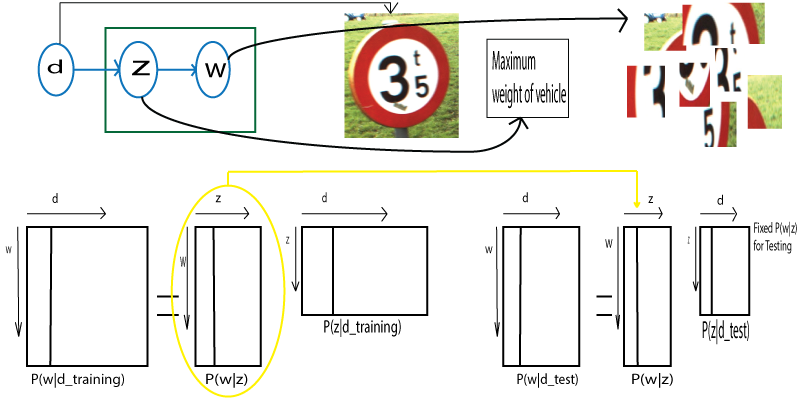}

  \caption{pLSA algorithm idea}

  \label{AAAa}

\end{figure*}

\subsubsection{LLC Codebook Formation}
For the formation of visual words locality constrained linear coding [18] based method is used. This algorithm generate similar codes for similar descriptors by sharing bases. Locality also leads to sparsity. Idea based locality importance more than sparsity is used and given by below optimization problem. Here X is a D dimensional local descriptors, $ X = [x_{1},x_{2},...,x_{N}] \in \Re^{DxN}$ and $ B = [b_{1},b_{2},...,b_{M}] \in \Re^{DXM}$ is M dimensional codebook.
\begin{equation}
min_{c}\Sigma_{i=1,N}||x_{i} - Bc_{i}||^{2} + \lambda||d_{i} \odot c_{i}||^{2}
\end{equation}
\begin{equation}
s.t. 1^{T}c_{i} = 1, \forall i
\end{equation}
\begin{equation}
d_{i} = exp(\frac{dist(x_{i},B)}{\sigma})
\end{equation}
where $\odot $ denotes the element-wise multiplication, and $d_{i} \in \Re^{M}$ is the locality adaptor that gives different freedom for each basis vector proportional to its similarity to the input descriptor$ x_{i} $. Also $ dist(x_{i},B) $ is the Euclidean distance between $x_{i}$  and B. Shift invariance nature is confirmed by the constraint equation (3). In our case we will have 128 dimension SIFT descriptors $x_{i}$ for each image. And we will form a codebook B of size 300 words. Each image will be expressed as a combination of these words. Each image will be represented as histogram of visual words. In Fig. [4] codebook formation idea is depicted.
\begin{figure}
  \centering
      \includegraphics[width=3.5in,height=2.4in]{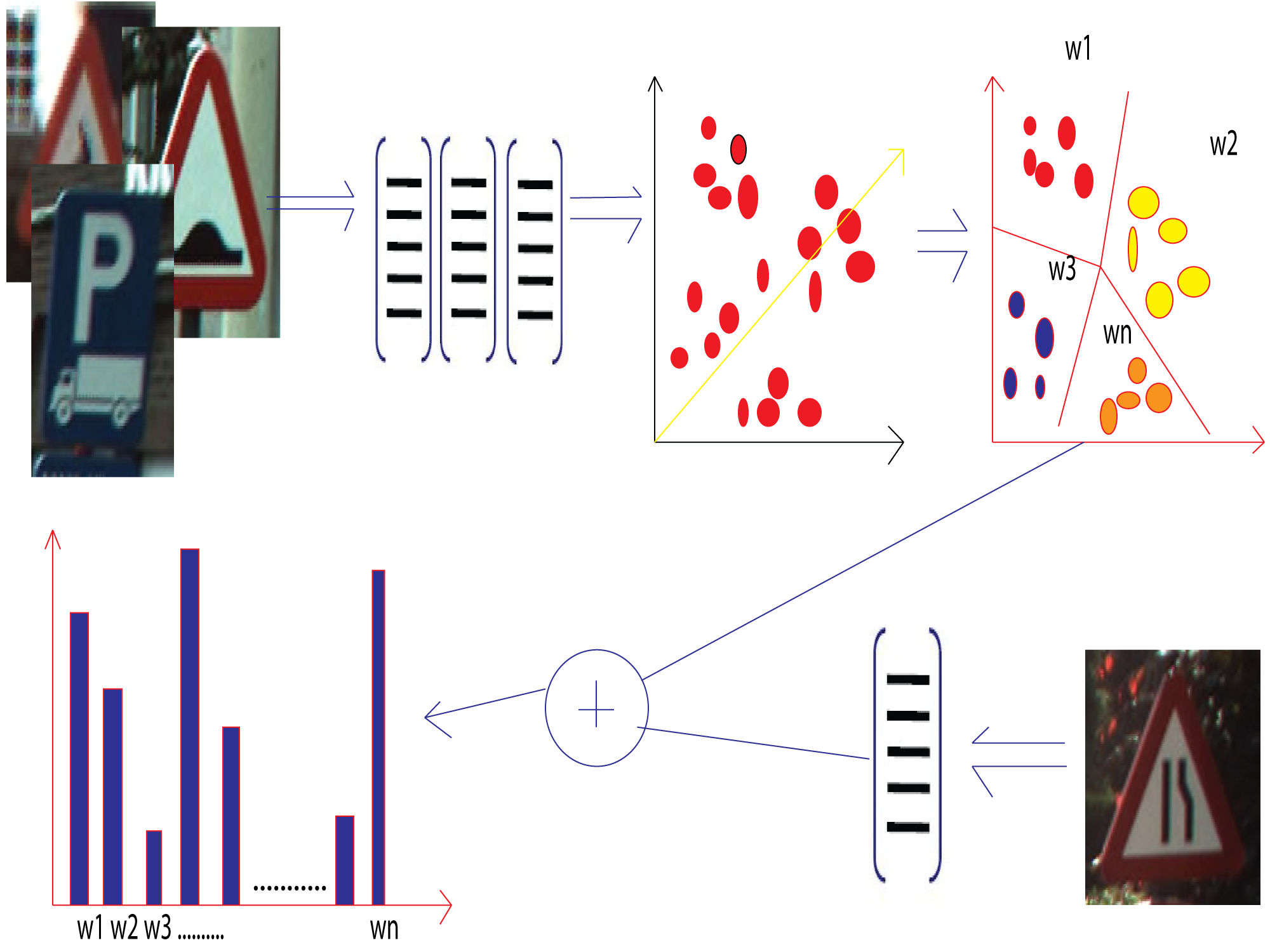}
\caption{Codeword formation}
\end{figure}

\begin{figure}
  \centering
      \includegraphics[width=3.2in,height=2.3in]{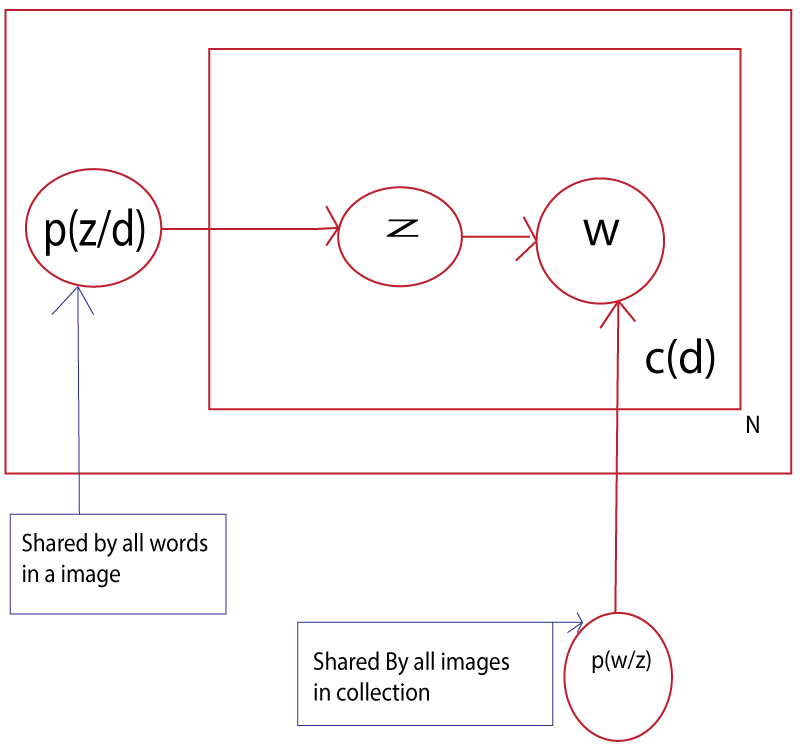}
\caption{pLSA concept}
\end{figure}

\subsubsection{pLSA model}
Probabilistic latent semantic analysis [7] is a topic discovery model, its concept based on latent variable analysis. An image also can be considered as a collection of topics. Every image can be considered as a text document with words from a specific vocabulary. The vocabulary will be formed by using k-means algorithm on SIFT features from training image. Suppose we have a collection of N images(document) $ D ={d_{1},d_{2},...,d_{N}}$, and corresponding vocabulary  with size N1 is $W={w_{1},w_{2},...,w_{N1}}$. And let there be N2 topic $Z={z_{1},z_{2},...,z_{N2}}$ Model parameter are computed using expectation maximization method.
In Fig. [3] and Fig. [5] pLSA algorithm idea and typical scenario is explained.

\begin{equation}
P(z|d,w) = \frac{P(z)P(d|z)P(w|z)}{\Sigma_{z'} P(z')P(d|z')P(w|z')}
\end{equation}
\begin{equation}
P(w|z) = \frac{\Sigma_{d}n(d,w)P(z|d,w)}{\Sigma_{d,w'}n(d,w')P(z|d,w')}
\end{equation}
\begin{equation}
P(d|z) = \frac{\Sigma_{w}n(d,w)P(z|d,w)}{\Sigma_{d',w}n(d',w)P(z|d',w)}
\end{equation}
\begin{equation}
P(z) = \frac{\Sigma_{d,w}n(d,w)P(z|d,w)}{R}
\end{equation}

\begin{equation}
R \equiv\Sigma_{d,w}n(d,w)
\end{equation}

By using histogram of words concept, each image will be converted to a document with previously designed vocabulary.

\subsubsection{Fuzzy-KNN classification}
Since our framework based on topic modelling and discovery, fuzzy KNN classification [19] techniques is used for getting appropriate topic af a test image corresponding to training images. Fuzzy kNN perform better than traditional KNN algorithm, because it depends on weight of neighbours.
In the training stage we will calculate $P(w|z)$, which will be used as input for testing algorithm to compute $P(z|d_{test})$.
After that a K- nearest neighbour algorithm is used to classify these image by using probability distribution $P(z|d_{train})$.
\begin{equation}
u_{i}(x) = \frac{\displaystyle\sum\limits_{j=1,K}u_{ij}(\frac{1}{||x - x_{j}||^{\frac{2}{(m-1)}}})}{\displaystyle\sum\limits_{j=1,K}(\frac{1}{||x - x_{j}||^{\frac{2}{(m-1)}}})}
\end{equation}
$u_{i}$ is value for membership strength to be computed, and $u_{ij}$ is previously labelled value of i th class for j th vector.
Final class label of the query point x is computed as follows:

\begin{equation}
u_{0}(x) = arg max_{i}(u_{i}(x) )
\end{equation}

\begin{table*}[t]
  \centering
  \begin{tabular}{*{20}{c}}
\hline
 Algorithm & Speed Limits & Prohibitions & Derestrictions & Mandatory & Danger & Unique\\
\hline
Committee of CNNs[3] & 99.47 & 99.93 & 99.72 &99.89& 99.07 & 99.22\\
\hline
Human & 98.32 & 99.87 & 98.89 & 100.00 & 99.21 & 100.00\\
\hline
Multi-Scale CNN[15] & 98.61 & 99.87 & 94.44 & 97.18 & 98.3 & 98.63\\
\hline
\textbf{pLSA} & 98.82 & 98.27 & 97.93 & 96.86 & 96.95 & 100.00\\
\hline
Random Forest[16] & 95.95 & 99.13 & 87.50 & 99.27 & 92.08 & 98.73\\
\hline
LDA on HOG2 [2] & 95.37 & 96.80 & 85.83 & 97.18 & 93.73 & 98.63\\
\hline
 \end{tabular}
  \caption{}
\end{table*}

\section{ Experiment and Result}
For testing accuracy of our algorithm we have used well known, German Traffic Sign Recognition Benchmark (GTSRB) [17], having 12000 training image of different class, it has 43 different class of signs. This dataset contain wide variety of image with the consideration of different challenge may involve in recognition. We have also used Belgium traffic sign recognition dataset [5] and used extra 19 classes of sign which were not included in GTSRB. This classes mainly related to different warning.  Different deformation due to viewpoint variation, occlusion due to different obstacles like trees, building etc., natural degrading, different weather condition are considered in these dataset. Also for training the system we have used different version of same images, specifically rotation of a image approximately $4\,^{\circ}{\rm C}$ in left and right direction also translated version of original image. In addition to that images are prepossed using contrast adjustment method, both the original and contrast adjusted images are used for training. Total number of images per category is shown in Table. [2]. 
\begin{table}[h]
\caption{}
\label{table_example}
\begin{center}
\begin{tabular}{|c|c|c|}
\hline
 Training set & Validation set & Test set\\
\hline
300 & 150 & 100\\
\hline

\end{tabular}
\end{center}
\end{table}

\begin{table}[h]
\caption{}
\label{table_example}
\begin{center}
\begin{tabular}{|c|c|}
\hline
Algorithm & Accuracy(\%)\\
\hline
Committe of CNNs & 99.46\\
\hline
Human Performance & 98.84\\
\hline
Multi-Scale CNNs & 98.31\\
\hline
\textbf{pLSA} & 98.14\\
\hline
Random Forest & 96.14\\
\hline
LDA on HOG2 & 95.68\\
\hline
LDA on HOG1 & 93.18\\
\hline
LDA on HOG3 & 92.34\\
\hline
\end{tabular}
\end{center}
\end{table}

\begin{figure}
  \centering
      \includegraphics[width=3.4in,height=2.0in]{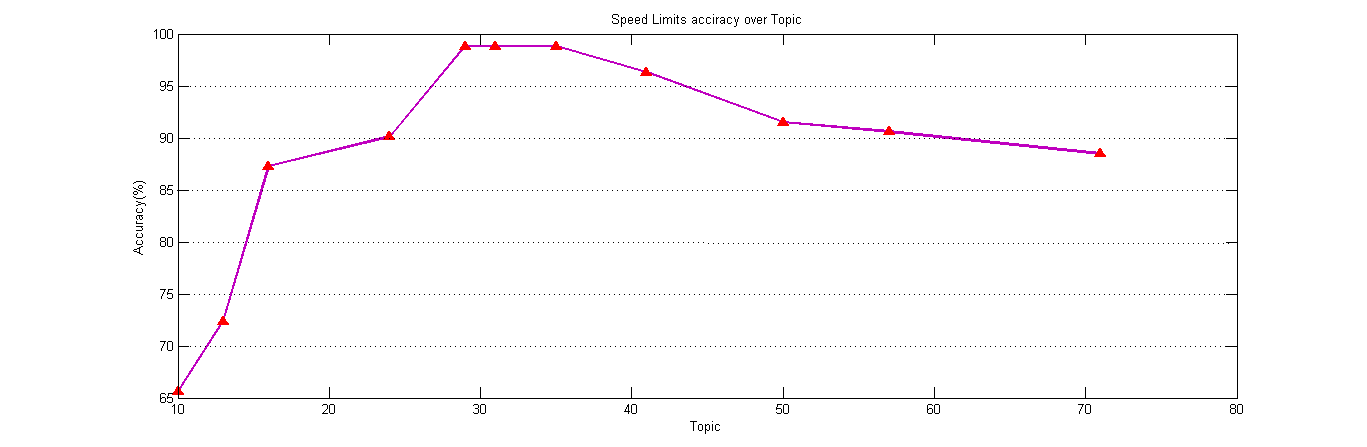}
\caption{Accuracy of Speed Limits category over TOPICs}
\end{figure}

Before running main algorithm on these image we will resize our image to $90 \times 97$. For testing purpose we will use affine transformation of same image to produce different version of it for getting rid of viewpoint variation. plSA is an iterative algorithm, for convergence of parameters we have used 182 iterations. In Fig. [7] we have shown log-likelihood variation with respect to number of iteration.

\begin{figure}
  \centering
      \includegraphics[width=3.3in,height=2.0in]{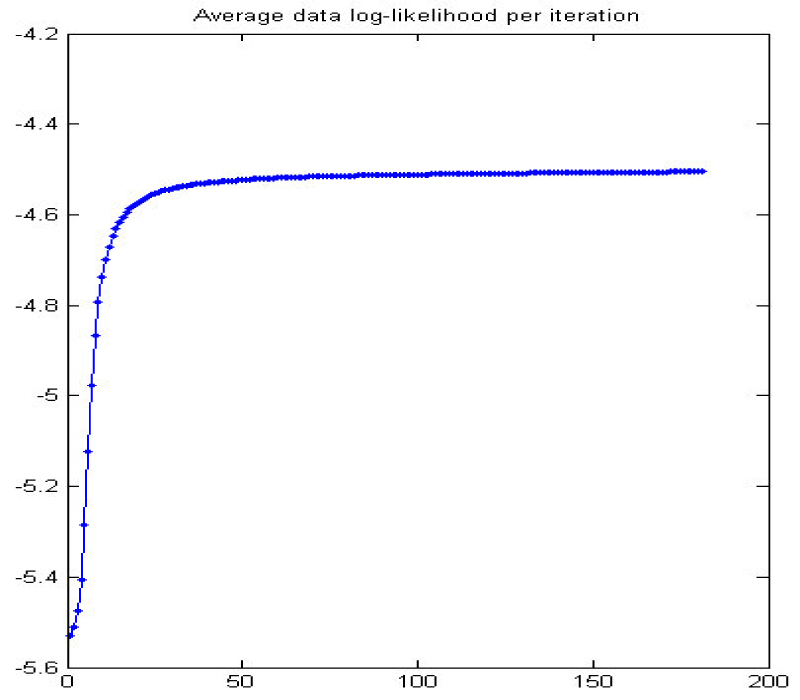}
\caption{Log-likelihood over iterations}
\end{figure}

First these image are classified on the basis of their shape by using HOG features. Once we have the prior information about the shape, we have applied our classification algorithm which is based on sift features.
All testing image will be pre-processed and converted to combination of visual words using LLC codebook formation method as described in earlier section for training images.
Our method final result is very promising. A comparison with different state of art methods is shown in Table.[3]. Also we will compare accuracy of some sub classes, like speed limits, other prohibitions, derestrictions, danger, mandatory and priority. We have almost very good classification result across sub classes. In Table.[1] a comparison of results obtained by different methods is explored.

The algorithm is implemented in a MATLAB platform also with some C++ rappers. All the testing was carried out in a quad core intel i7 Linux Machine. 

Performance of our algorithm vary with numbers of topic used in classification. In Fig. [6] we have shown a comparison of accuracy over topics for speed limits subclass. Similar accuracy variation over topics is obtained for other subcategory. We have maximum accuracy over 29-44 number of topics. We have chosen 35 number of topics as optimal for classification of our system.

\section{Conclusions}
In this work we have developed a novel and fast approach for traffic sign recognition problems. Because of little variability between different same shaped sign it is very difficult problems in classification. We exploit the concept of representation of images as collection of words and topic, by using the visual codebook formation concepts. For computation of probability of topics in images we have used probabilistic latent semantic analysis. Our main contribution is using novel pLSA model for getting near state of art performance in classification problems and developing a shape based classifier. This system is independent of colour. Hardware implementation of this approach is much easier than other SVM, neural network and deep learning based approach.

% The following two commands are all you need in the
% initial runs of your .tex file to
% produce the bibliography for the citations in your paper.

% that's all folks
\end{document}